\documentclass[3p,,preprint,12pt]{elsarticle}
\makeatletter\if@twocolumn\PassOptionsToPackage{switch}{lineno}\else\fi\makeatother

\usepackage{tabulary,xcolor}
\usepackage{amsfonts,amsmath,amssymb}
\usepackage[T1]{fontenc}
\makeatletter
\let\save@ps@pprintTitle\ps@pprintTitle
\def\ps@pprintTitle{\save@ps@pprintTitle\gdef\@oddfoot{\footnotesize\itshape \null\hfill\today}}
\def\hlinewd#1{%
  \noalign{\ifnum0=`}\fi\hrule \@height #1%
  \futurelet\reserved@a\@xhline}
\def\tbltoprule{\hlinewd{.8pt}\\[-12pt]}
\def\tblbottomrule{\noalign{\vspace*{6pt}}\hline\noalign{\vspace*{2pt}}}
\def\tblmidrule{\noalign{\vspace*{6pt}}\hline\noalign{\vspace*{2pt}}}

\AtBeginDocument{\ifNAT@numbers \biboptions{sort&compress}\fi}

\makeatother

\usepackage{ifluatex}
\ifluatex
\usepackage{fontspec}
\defaultfontfeatures{Ligatures=TeX}
\usepackage[]{unicode-math}
\unimathsetup{math-style=TeX}
\else 
\usepackage[utf8]{inputenc}
\fi 
\ifluatex\else\usepackage{stmaryrd}\fi

\usepackage{url,multirow,morefloats,floatflt,cancel,tfrupee}
\makeatletter

\AtBeginDocument{\@ifpackageloaded{textcomp}{}{\usepackage{textcomp}}}
\makeatother
\usepackage{colortbl}
\usepackage{xcolor}
\usepackage{pifont}
\usepackage[nointegrals]{wasysym}
\makeatletter

\def\mcWidth#1{\csname TY@F#1\endcsname+\tabcolsep}

\def\cAlignHack{\rightskip\@flushglue\leftskip\@flushglue\parindent\z@\parfillskip\z@skip}
\def\rAlignHack{\rightskip\z@skip\leftskip\@flushglue \parindent\z@\parfillskip\z@skip}

\@ifundefined{etal}{}{}

\usepackage{ifxetex}
\ifxetex\else\if@twocolumn\@ifpackageloaded{stfloats}{}{\usepackage{dblfloatfix}}\fi\fi

\AtBeginDocument{
\expandafter\ifx\csname eqalign\endcsname\relax
\def\eqalign#1{\null\vcenter{\def\\{\cr}\openup\jot\m@th
  \ialign{\strut$\displaystyle{##}$\hfil&$\displaystyle{{}##}$\hfil
      \crcr#1\crcr}}\,}
\fi
}

\AtBeginDocument{%
  \@ifpackageloaded{endfloat}%
   {\renewcommand\efloat@iwrite[1]{\immediate\expandafter\protected@write\csname efloat@post#1\endcsname{}}}{\newif\ifefloat@tables}%
}%

\def\BreakURLText#1{\@tfor\brk@tempa:=#1\do{\brk@tempa\hskip0pt}}
\let\lt=<
\let\gt=>
\def\processVert{\ifmmode|\else\textbar\fi}

\@ifundefined{subparagraph}{
\def\subparagraph{\@startsection{paragraph}{5}{2\parindent}{0ex plus 0.1ex minus 0.1ex}%
{0ex}{\normalfont\small\itshape}}%
}{}

\newcommand\role[1]{\unskip}
\newcommand\aucollab[1]{\unskip}
  
\@ifundefined{tsGraphicsScaleX}{\gdef\tsGraphicsScaleX{1}}{}
\@ifundefined{tsGraphicsScaleY}{\gdef\tsGraphicsScaleY{.9}}{}
\def\checkGraphicsWidth{\ifdim\Gin@nat@width>\linewidth
	\tsGraphicsScaleX\linewidth\else\Gin@nat@width\fi}

\def\checkGraphicsHeight{\ifdim\Gin@nat@height>.9\textheight
	\tsGraphicsScaleY\textheight\else\Gin@nat@height\fi}

\def\fixFloatSize#1{}
\let\ts@includegraphics\includegraphics

\def\inlinegraphic[#1]#2{{\edef\@tempa{#1}\edef\baseline@shift{\ifx\@tempa\@empty0\else#1\fi}\edef\tempZ{\the\numexpr(\numexpr(\baseline@shift*\f@size/100))}\protect\raisebox{\tempZ pt}{\ts@includegraphics{#2}}}}

\AtBeginDocument{\def\includegraphics{\@ifnextchar[{\ts@includegraphics}{\ts@includegraphics[width=\checkGraphicsWidth,height=\checkGraphicsHeight,keepaspectratio]}}}

\DeclareMathAlphabet{\mathpzc}{OT1}{pzc}{m}{it}

\def\URL#1#2{\@ifundefined{href}{#2}{\href{#1}{#2}}}

\def\UrlOrds{\do\*\do\-\do\~\do\'\do\"\do\-}%
\g@addto@macro{\UrlBreaks}{\UrlOrds}

\edef\fntEncoding{\f@encoding}

\makeatother

\newif\ifmultipleabstract\multipleabstractfalse%
    \makeatletter
\def\ead{\@ifnextchar[{\@uad}{\@ead}}
\gdef\@ead#1{\bgroup
   \def\_{\string\underscorechar\space}
   \def\{{\string\lbracechar\space}
   \def\textdagger{\string\textdagger\space}
   \def\texttildeapprox{\string\texttildeapprox\space}
   \def~{\hashchar\space}
   \def\}{\string\rbracechar\space}
   \edef\tmp{\the\@eadauthor}
   \immediate\write\@auxout{\string\emailauthor
     {#1}{\expandafter\strip@prefix\meaning\tmp}}
  \egroup
}
\gdef\emailauthor#1#2{\stepcounter{ead}
      \g@addto@macro\@elseads{\raggedright
      \let\corref\@gobble
      \eadsep\texttt{#1} (#2)
      \def\eadsep{\unskip,\space}}
}

\makeatother
  
\let\citep\cite
\let\citet\cite    
  
\usepackage{float}

\begin{document}

\begin{frontmatter}

    \title{
  Transformer Reconstructed with Dynamic Value Attention    
}
    
\author[]{Xiaowei Wang\corref{c-b27ad86d2907}}
\ead{wangxiaowei@cup.edu.cn}\cortext[c-b27ad86d2907]{Corresponding author.}
    
\address{College of Artificial Intelligence\unskip, 
     China University of Petroleum (Beijing)\unskip, Beijing\unskip, 102249\unskip, China}

\begin{abstract}
Since transformer was firstly published in 2017, several works have been proposed to optimize it. However, the major structure of transformer remains unchanged, ignoring one of its main intrinsic limitations, which is the same static value is used for every query in a head. Transformer itself tries to solve this problem by implementing multi-head attentions, yet the number of heads is limited by complexity. I~propose a method to decide a value for each query dynamically, which could cut down all the redundant heads, keeping only one. Consequently, the following feed forward network could be cut down entirely, as each revised embedding has already fetched enough useful values far beyond the context. As a result, a single-head Dynamic Value Attention (DVA) is all you need in a transformer. According to the experiment, DVA may save 37.6\% training time than the original transformer meanwhile increasing the learning capability.
\end{abstract}
    
  \end{frontmatter}

\section{Introduction}
In Natural Language Processing (NLP) domain, word2vec\unskip~\cite{2850382:34216485}  tries to encode each token (a word or a piece of word) into a vector, also called embedding, and an embedding is an efficient media to represent the semantic knowledge of the token, and this lays the foundation for the current brilliant NPL applications. Based on this form of knowledge representation, Machine Learning (ML) technologies, especially neural network models, are used to process NLP tasks. In a neural network model, the input would be transmitted and manipulated along the hidden layers of a network, and the most exciting thing is that the network could learn the right way to do its job by studying historical examples automatically. 

Traditional neural networks only accept fixed length inputs, and to deal with unfixed length inputs, Recurrent Neural Network (RNN) \unskip~\cite{2850382:34216498} was proposed. Simply put, RNN fetch the semantic knowledge of the input embeddings one by one, and each embedding's knowledge will be melted into a unique hidden layer. Finally, when the knowledge of all the preceding input embeddings are melted into the hidden layer, the hidden layer now could be used to predict the final output. Although RNN is a pioneering method for handling unfixed length input embeddings, it is hard to train in practice. On one hand, it fetches fixed rate knowledge from each embedding, and this decays the learning capability of the hidden layer. On the other hand, it is very hard to train because of the vanishing and exploding gradient problems, and its non-parallel working method. 

Linear RNN\unskip~\cite{2850382:34216499} was proposed to tackle the training problems of RNN, which reinforces RNN with parallel computation without gradient vanishing or exploding problems. It works well on very long-range dependencies, which means it could support very large context size \unskip~\cite{2850382:34216536}. However, linear RNN doesn't work well in NLP tasks, because it pays the same attention to every input embedding, which makes irrelevant knowledge accumulates in the hidden layer. Long Short-Term Memory (LSTM)\unskip~\cite{2850382:34216537}, Mamba \unskip~\cite{2850382:34216538} and Sequence to Sequence (seq2seq)\unskip~\cite{2850382:34216578}  try to figure out when to store the knowledge and when to forget the knowledge based on what input they see. LSTM and Mamba only pay attention to individual input embeddings separately, in other words, each input embedding always provides the same knowledge without taking care of relationships between embeddings. 

The self-attention mechanism in transformer\unskip~\cite{2850382:34203869} does take care of the strength relationships between input embeddings, by calculating a Scaled Dot Product Attention(SDPA) weight matrix, and each score in the SDPA represents the relation strength between the two corresponding input embeddings . The self-attention mechanism provides an efficient way to represent and manipulate semantic knowledge in the form of embeddings, by paying specific attention to the relationships between them. This choice makes it not only works well for NLP tasks, but also does good jobs in other domains, as any form of knowledge representation could be converted into the form embeddings. For example, a word of cat may share the same embedding with a picture of cat. As a result, it has become a core Artificial General Intelligence (AGI) component of almost all mainstream large-scale models, and multimodal large language models also become very popular.

To make a further step, I~try to reconstruct the transformer by calculating not only the strength relationships between embeddings, but also measuring the semantic relationships between embeddings. This is done by adopting a mechanism to dynamically produce a value embedding for each pair of query embedding and key embedding. Although, it looks like additional work has been implemented, but with this Dynamic Value Attention (DAV) mechanism, only one-head transformer is needed without any feed forward network block. According to the experiment, training time could be cut down dramatically. Also, as DAV mechanism pays attention both to the strength relationship and semantic relationship between embeddings, and it shows better learning capability during the experiment.
    
\section{Baseline}
I take the original transformer model as my baseline in this paper, which contains two main blocks including a multi-head attention block and a feed forward network block. To better convey the idea of DVA, a semantic explanation of transformer is presented as follows, with which it might be easier to explain why a Large Language Models (LLM) may exhibit human-like intelligence.

\subsection{Single-head attention block explained}In a single-head attention block of transformer, the relationships between embeddings are recorded in a score matrix similar in seq2seq, but the scores are calculated in a more sophisticated way. In seq2seq, dot-product serves to evaluate the similarity, but not the strength relationships between embedding vectors. To get more accurate descriptions, we might need to create a new strength evaluator to assess the strength relationships between embedding vectors, and that is exactly what the attention block does in transformer. 
\let\saveeqnno\theequation
\let\savefrac\frac
\def\dispfrac{\displaystyle\savefrac}
\begin{eqnarray}
\let\frac\dispfrac
\gdef\theequation{1}
\let\theHequation\theequation
\label{dfg-40cd38ee09fd}
\begin{array}{@{}l}s(B,A)=q(B)\cdot k(A)=B\times W_Q\cdot A\times W_K=query_B\cdot key_A\end{array}
\end{eqnarray}
\global\let\theequation\saveeqnno
\addtocounter{equation}{-1}\ignorespaces 
For example, let's assume vector B is querying on vector A, and Equation~(\ref{dfg-40cd38ee09fd}) might be proposed. Firstly, we want to measure the strength relation between B and A, which is s(B,A). Then, we can expand it into the dot-product of q(B) and k(A). In other words, the proper choice of q() and k() could leverage the dot-product q(B) and k(A) to represent the strength relation between B and A, which is s(B, A). It is hard for human to craft such functions, but we can explore the power of neural network to urge W\ensuremath{_{Q}} and W\ensuremath{_{K}} to mimic q() and k() by letting $\;q(B)=B\times W_Q,\;k(A)=A\times W_K $, and fortunately W\ensuremath{_{Q\ }}and W\ensuremath{_{K}} could be tuned to the proper status through training processes. These two functions q() and k() should not be interpreted separately, they are always working together, making their dot-product meaningful to assess the strength relationships between related embedding vectors. As a result, the dot-product of query\ensuremath{_{B}} and key\ensuremath{_{A}} could represent the strength relation between B and A.
\let\saveeqnno\theequation
\let\savefrac\frac
\def\dispfrac{\displaystyle\savefrac}
\begin{eqnarray}
\let\frac\dispfrac
\gdef\theequation{2}
\let\theHequation\theequation
\label{dfg-d845dd0ef6f9}
\begin{array}{@{}l}v(A)=A\times W_V=value_A\end{array}
\end{eqnarray}
\global\let\theequation\saveeqnno
\addtocounter{equation}{-1}\ignorespaces 
Here comes value\ensuremath{_{A}}, which should contain A's contributing semantic value to queries. If we craft a function v(), fulfilling v(A) =value\ensuremath{_{A}}, v() could be interpreted as a function to capture the influence which A could provide to any query. So that, value\ensuremath{_{A}} represents the A's influence on B and other embeddings. Again, such v() is hard to build by human being, but we can rely on the power of neural network to urge W\ensuremath{_{V}} to mimic v() by letting $v(A)\;=\;A\times W_V $, and W\ensuremath{_{V}} could be tuned during training processes. Finally, if no other embedding exists, Vector B may achieve its refined version B' as $B'=query_B\cdot key_A\times value_A+query_B\cdot key_B\times value_B $, which means $B'=\;Score_{BA}\times value_A+Score_{BB}\times value_B $.

\subsection{Multi-head attention block explained}Let's check this sentence, \textit{hot girl drinks hot water.} So, what does \textit{hot }mean then, lovely or heated? We human beings know that the first hot means lovely for the girl and the second hot means heated for the water, yet according to single-head attention mechanism, the two hots could only share one meaning. That is because only one value function v() powered by the weight matrix W\ensuremath{_{V}} is available in a single head, and the result value of v(hot) is fixed in the head. This mechanism prevents the querying embedding vectors (girl and water in this example) from fetching appropriate value updates (value contributed by hot in this example).

Multi-head attention mechanism is adopted to mitigate this problem. For example, in Figure~\ref{f-183ff6386284}, head1 and head2 are implemented for two different situations. In head1, v(hot)=lovely, and the model could adjust the score matrix to make sure only the girl gets the value of lovely in this head. In head2, v(hot)=heated, and the model again could adjust the score matrix to make sure only the water gets the value of heated in this head. At the end, head1 and head2 could be combined together to ensure that both the girl and the water get their appropriate value updates.

\bgroup
\fixFloatSize{images/3638ba1c-8643-4721-92ea-47992ab4f00f-ufigure1-u-value-contribution-in-multi-head-attention.png}
\begin{figure*}[!htbp]
\centering \makeatletter\IfFileExists{images/3638ba1c-8643-4721-92ea-47992ab4f00f-ufigure1-u-value-contribution-in-multi-head-attention.png}{\includegraphics{images/3638ba1c-8643-4721-92ea-47992ab4f00f-ufigure1-u-value-contribution-in-multi-head-attention.png}}{\includegraphics{3638ba1c-8643-4721-92ea-47992ab4f00f-ufigure1-u-value-contribution-in-multi-head-attention.png}}
\makeatother 
\caption{{Value contribution in multi-head attention}}
\label{f-183ff6386284}
\end{figure*}
\egroup
For sure, multi-head attention mechanism solves this problem to some extends, but it is limited by the number of available heads which is restricted by computation resources. DVA tries to figure out this problem in only one head, which means each keying embedding vector (hot in this example) may provide different value updates for different querying embedding vectors (girl and water in this example). In other words, each pair of key and query dynamically defines a specialized value update for the querying embedding vector.

\subsection{Feed forward network block explained}In transformer, multi-head attention adjusts a querying embedding vector's meaning in the context, or more specifically in each head a querying embedding vector's meaning is updated by all keying embedding vectors in the context, and then the meaning updates for the querying embedding vector in all heads are combined together.

Different from multi-head attention block, the Feed Forward Network (FFN) block tries to adjust each individual querying embedding vector by a broader range of possible value updates outside the context. A FFN block is usually interpreted as a simple Multilayer Perceptron (MLP), through which an embedding in the dimension of n is firstly projected into a hidden layer in a higher dimension of $\alpha n $, and then projected back into a refined embedding in the dimension of n, the same as the original input embedding. However, this view of interpretation leaves the refining process of an embedding as a black box, and a clearer semantic explanation is deserved as shown in Figure~\ref{f-05541b20514d}.

\bgroup
\fixFloatSize{images/36d7695e-1d2c-4440-8cc3-091ff50493c6-ufigure2-u-embedding-adjustment-in-feed-forward-network-block.png}
\begin{figure*}[!htbp]
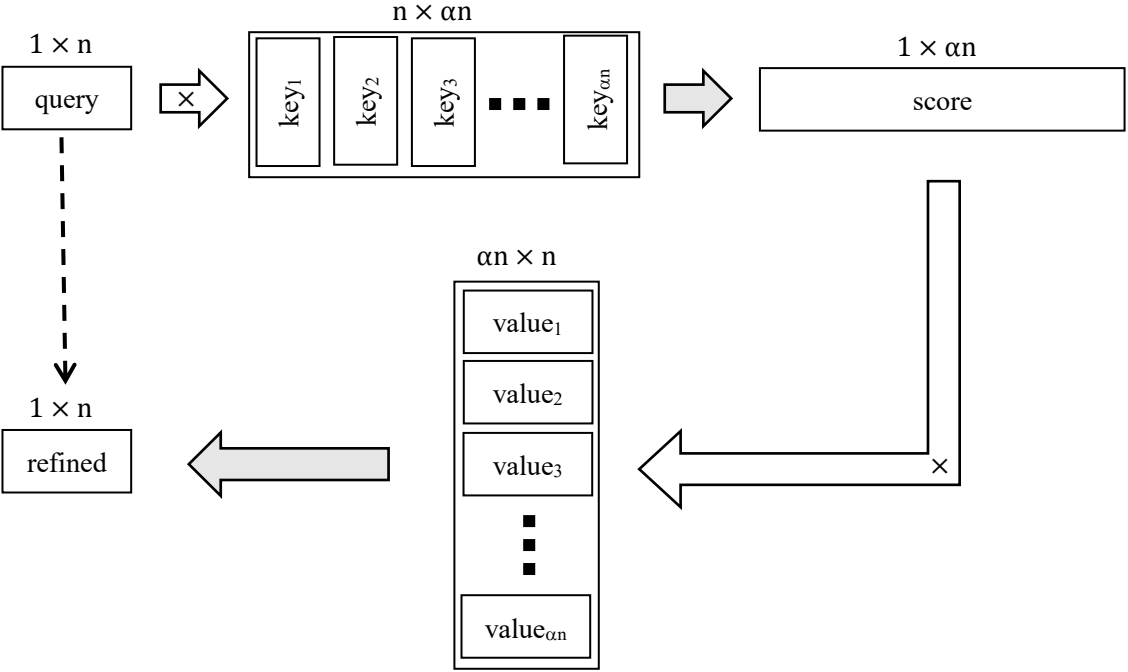

\centering \makeatletter\IfFileExists{images/36d7695e-1d2c-4440-8cc3-091ff50493c6-ufigure2-u-embedding-adjustment-in-feed-forward-network-block.png}{\includegraphics{images/36d7695e-1d2c-4440-8cc3-091ff50493c6-ufigure2-u-embedding-adjustment-in-feed-forward-network-block.png}}{\includegraphics{36d7695e-1d2c-4440-8cc3-091ff50493c6-ufigure2-u-embedding-adjustment-in-feed-forward-network-block.png}}
\makeatother 
\caption{{Embedding adjustment in feed forward network block}}
\label{f-05541b20514d}
\end{figure*}
\egroup
An input embedding could be interpreted as a query resulting from a function q(embedding), yet q() works as a simple unit matrix unchanging the value of the input embedding, and this assumption enables us to recognize the process of FFN as a cross-attention process. Under the perspective of this view, the upper weight matrix works as a list of keys, and the lower weight matrix works as a list of values. The query pays attentions to the keys according to the score vector, as the score vector is derived from the dot-product of each pair of query and key. Then the multiplication of the score and values indicates each value update the querying embedding vector gets from the corresponding keying embedding vectors, and the combination of them is the total value update for the querying embedding vector.

However, where do the keys and values come from then? A possible explanation is that they come from a latent Core Embedding Set (CES), which contains the most important embeddings believed by the transformer through training processes. Each key could be derived from the latent function k(embedding\ensuremath{_{CES}}), and also each value could be derived from the latent function v(embedding\ensuremath{_{CES}}), and every embedding\ensuremath{_{CES}} comes from the CES. In other words, with the CES and k() and v(),  both the upper weight matrix as keys and the lower weight matrix as values could be derived, and the CES and k() and v() are latent trained weights. As the CES looks like a latent embedding dictionary of an unknown language, it seems an input embedding coming from the context pays cross-attention to all the embeddings in CES, and the CES acts as a knowledge base to some extends.

Although the FFN block enables an embedding to get richer semantic meanings through a cross-attention like process, it is limited in two aspects according to the structure of FFN. Firstly, only single-head cross-attention is available, so that the value contributions from CES are constrained. Then, the function q() doesn't work, only the function k() works, and this may decay the scoring quality based on the dot-product of q() and k(). However, with DVA, no FFN is needed anymore. Because each querying embedding vector may have already got enough value updates in a single-head attention block. In DVA, possible values are derived dynamically from the semantic relationships between pairs of query and key, which means DVA could provide possible value updates far beyond the context.
    
\section{Model Principles}
The core of DVA is to craft a method to assess the semantic relationship between two embeddings. As mentioned ahead, in Equation~(\ref{dfg-40cd38ee09fd}) ,  the strength relationship s(B, A) is captured by the dot-product of q(B) and k(A), and this is enabled by the trainable weight matrices W\ensuremath{_{Q}} and W\ensuremath{_{K}}. According to this idea, the semantic relationship r(B, A) might also be captured by some calculation between q\ensuremath{_{R}}(B) and k\ensuremath{_{R}}(A), and this could also be enabled by some trainable weight matrices W\ensuremath{_{QR}} and W\ensuremath{_{KR}}, and two possible solutions are proposed as follows.

\subsection{Initial Solution}Firstly, a straight-forward solution is proposed through Equation~(\ref{dfg-165ce730d69f}), Equation~(\ref{dfg-83a41e362825}), Equation~(\ref{dfg-f3c6fbd0afb7}). The semantic relationship between the querying embedding vector B and the keying embedding vector A could be measured by r(B, A), and this is enabled by the proper choice of the functions q\ensuremath{_{R}}() and k\ensuremath{_{R}}(), which could leverage the matrix multiplication of q\ensuremath{_{R}}(B)\ensuremath{^{T}} and k\ensuremath{_{R}}(A)\ensuremath{^{}}as a linear transformation matrix. Again, it is hard for human to craft such functions, but we can explore the power of neural network to urge W\ensuremath{_{QR\ }}and W\ensuremath{_{KR\ }}to mimic q\ensuremath{_{R}}() and k\ensuremath{_{R}}() during training processes.
\let\saveeqnno\theequation
\let\savefrac\frac
\def\dispfrac{\displaystyle\savefrac}
\begin{eqnarray}
\let\frac\dispfrac
\gdef\theequation{3}
\let\theHequation\theequation
\label{dfg-165ce730d69f}
\begin{array}{@{}l}r(B,\;A)=q_R(B)^{T}\times k_R(A)=(B\times W_{QR})^{T}\times(A\times W_{KR})=query_{BR}^{T}\times key_{AR}\end{array}
\end{eqnarray}
\global\let\theequation\saveeqnno
\addtocounter{equation}{-1}\ignorespaces 

\let\saveeqnno\theequation
\let\savefrac\frac
\def\dispfrac{\displaystyle\savefrac}
\begin{eqnarray}
\let\frac\dispfrac
\gdef\theequation{4}
\let\theHequation\theequation
\label{dfg-83a41e362825}
\begin{array}{@{}l}v_R(B,\;A)=v(A)\;\times\;r(B,\;A)\end{array}
\end{eqnarray}
\global\let\theequation\saveeqnno
\addtocounter{equation}{-1}\ignorespaces 

\let\saveeqnno\theequation
\let\savefrac\frac
\def\dispfrac{\displaystyle\savefrac}
\begin{eqnarray}
\let\frac\dispfrac
\gdef\theequation{5}
\let\theHequation\theequation
\label{dfg-f3c6fbd0afb7}
\begin{array}{@{}l}v_{SR}(B,\;A)=s(B,\;A)\;\times\;v_R(B,\;A)\end{array}
\end{eqnarray}
\global\let\theequation\saveeqnno
\addtocounter{equation}{-1}\ignorespaces 
Anyway, r(B, A) might capture the semantic relationship between B and A, and a dynamic value update for B here could be derived from Equation~(\ref{dfg-83a41e362825}) as v\ensuremath{_{R}}(B, A), and it is the result of a linear transformation of v(A) by r(B, A). The r(B, A) acts as a modifier to v(A), dynamically fetching proper semantic knowledge from v(A) for the querying embedding vector B. Combining both the strength and semantic relationships between B and A, the final value update from A to B is captured in v\ensuremath{_{SR}}(B, A) through Equation~(\ref{dfg-f3c6fbd0afb7}), and it could be interpreted as a weighted semantic value attention.

According to this mechanism, a querying embedding vector could fetch different weighted semantic value updates from every keying embedding in the context. The most important thing is that v\ensuremath{_{R}}(B, A) is not fixed as v(A) in a single-head, but is dynamically decided by the semantic relationship between B and A. It seems a reasonable solution, but it costs too much computation resource, which makes it infeasible in practice. It is not a good solution initially, but some tiny changes could make the idea of DVA feasible, efficient and powerful shown as follows.

\subsection{Revised Solution}The modifier r(B, A) in Equation~(\ref{dfg-165ce730d69f}) for v(A) is in the form of a linear transformation matrix, which is the result of a matrix multiplication between query\ensuremath{_{BR}}\ensuremath{^{T}} and key\ensuremath{_{AR}}. The modifying process is another matrix multiplication between v(A) and r(B, A), so two extra matrix multiplications are needed for each pair of B and A. This costs too much, and a revised solution is proposed through Equation~(\ref{dfg-16c5e337bc80}), Equation~(\ref{dfg-1b30b409d2bd}), Equation~(\ref{dfg-4bee93239dbf}).
\let\saveeqnno\theequation
\let\savefrac\frac
\def\dispfrac{\displaystyle\savefrac}
\begin{eqnarray}
\let\frac\dispfrac
\gdef\theequation{6}
\let\theHequation\theequation
\label{dfg-16c5e337bc80}
\begin{array}{@{}l}r(B,\;A)=q_R(B)\;\ast\;k_R(A)=(B\times W_{QR})\;\ast\;(A\times W_{KR})\;=query_{BR}\ast key_{AR}\end{array}
\end{eqnarray}
\global\let\theequation\saveeqnno
\addtocounter{equation}{-1}\ignorespaces 

\let\saveeqnno\theequation
\let\savefrac\frac
\def\dispfrac{\displaystyle\savefrac}
\begin{eqnarray}
\let\frac\dispfrac
\gdef\theequation{7}
\let\theHequation\theequation
\label{dfg-1b30b409d2bd}
\begin{array}{@{}l}v_R(B,\;A)=v(A)\;+\;r(B,\;A)\end{array}
\end{eqnarray}
\global\let\theequation\saveeqnno
\addtocounter{equation}{-1}\ignorespaces 

\let\saveeqnno\theequation
\let\savefrac\frac
\def\dispfrac{\displaystyle\savefrac}
\begin{eqnarray}
\let\frac\dispfrac
\gdef\theequation{8}
\let\theHequation\theequation
\label{dfg-4bee93239dbf}
\begin{array}{@{}l}v_{SR}(B,\;A)=s(B,\;A)\times v_R(B,\;A)\end{array}
\end{eqnarray}
\global\let\theequation\saveeqnno
\addtocounter{equation}{-1}\ignorespaces 
In Equation~(\ref{dfg-16c5e337bc80}) , the modifier r(B, A) is represented by the broadcast decimal multiplication between q\ensuremath{_{R}}(B) and k\ensuremath{_{R}}(A), and the result is in the form of a embedding vector. As shown in Equation~(\ref{dfg-1b30b409d2bd}) , r(B, A), acting as a modifier, could be added to v(A) directly to produce a modified value update from A for B. So, the matrix multiplications are replaced by much simpler decimal multiplication and addition. Again, we can rely on the training processes to craft such functions q\ensuremath{_{R}}(B) and k\ensuremath{_{R}}(A), making whose decimal multiplication represent the semantic relationship between B and A, working as a modifier on v(A) for B. The revised solution shares the major idea of the initial solution, but implemented in a much simpler form, which makes it feasible, efficient and powerful.
    
\section{Training Experiment}
The experiment in this paper is based on a book \textit{Build a Large Language Model (From Scratch)} written by Sebastian Raschka \unskip~\cite{2850382:34228300} . It provides all the necessary components to build a LLM, and a standard Generative Pre-trained Transformer (GPT) model GPT-2 is chosen as a benchmark, as OpenAI has made the weights of GPT-2 publicly available. Bigger models, such as GPT-3, are not considered as the benchmark, because they are fundamentally the same in terms of model architecture, except they are scaled up to much bigger parameters. 

As shown in Figure~\ref{f-318bf67bf7b3} , part a in the left hand side represents the original GPT-2 model (model a), in which a transformer is composed of a masked multi-head attention block and a feed forward block. However, part b in the right hand side represents the reconstructed GPT-2 model (model b), in which a transformer is composed of a masked single-head DVA attention block only, without the following feed forward block. It is obvious in vision that model b is simpler and more compact than model a.

\bgroup
\fixFloatSize{images/b00eb703-599f-4f7a-8470-436844337a71-ufigure3-u-the-model-architectures-of-experiments.png}
\begin{figure*}[!htbp]
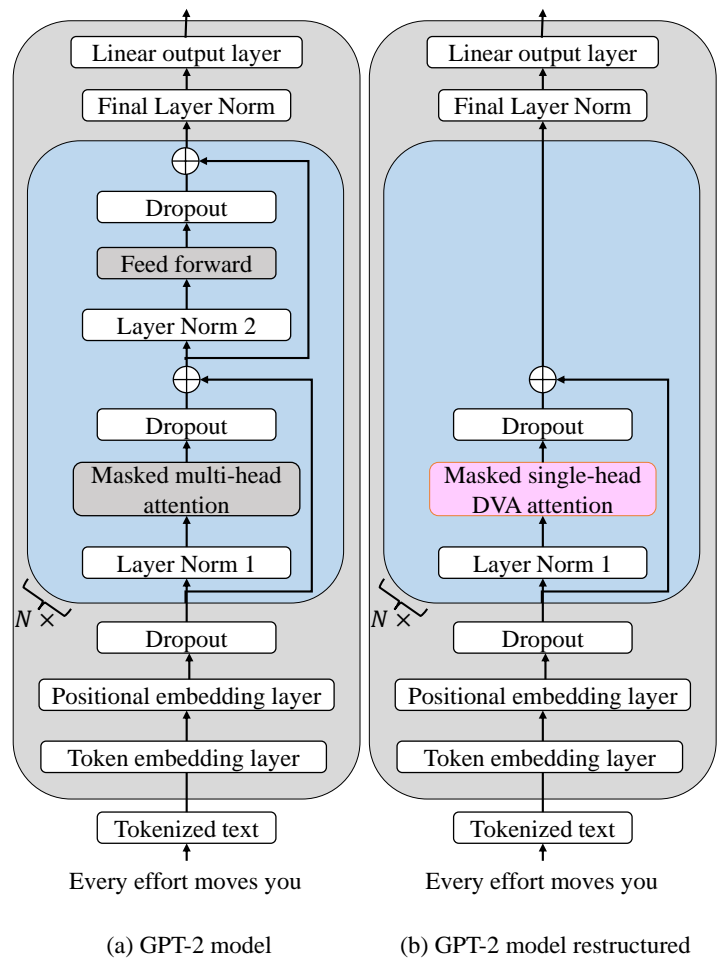

\centering \makeatletter\IfFileExists{images/b00eb703-599f-4f7a-8470-436844337a71-ufigure3-u-the-model-architectures-of-experiments.png}{\includegraphics[width=.59\linewidth]{images/b00eb703-599f-4f7a-8470-436844337a71-ufigure3-u-the-model-architectures-of-experiments.png}}{\includegraphics{b00eb703-599f-4f7a-8470-436844337a71-ufigure3-u-the-model-architectures-of-experiments.png}}
\makeatother 
\caption{{The architectures of the two models}}
\label{f-318bf67bf7b3}
\end{figure*}
\egroup
In more detail, the meta parameters of these two models are illustrated in Table~\ref{tw-888d1b4b4e1f} . Vocabulary size refers to a vocabulary of 50,257 words, as used by the tokenizer. Context length denotes the maximum number of input tokens the model can handle. Embedding dimension indicates transforming each token into a 768-dimensional embedding vector. Number of attention heads shows the count of attention heads, 12 in model a, and only 1 in model b. Number of layers specifies the number of transformer blocks in a model. Drop rate indicates the intensity of the dropout mechanism to prevent overfitting. Query-key-value bias determines whether to include a bias vector for query, key, and value computations. The last row of the table shows that model a has 162,419,712 parameters in total, and model b has 112,800,768 parameters in total.

\begin{table*}[!htbp]
\caption{{The meta parameters of the two models} }
\label{tw-888d1b4b4e1f}
\def\arraystretch{1}
\ignorespaces 
\centering 
\begin{tabulary}{\linewidth}{p{\dimexpr.3575\linewidth-2\tabcolsep}p{\dimexpr.30250000000000004\linewidth-2\tabcolsep}p{\dimexpr.34\linewidth-2\tabcolsep}}
\tbltoprule  & model a & model b\\
\tblmidrule 
vocabulary size &
  50257 &
  50257\\
context length &
  256 &
  256\\
embedding dimension &
  768 &
  768\\
number of attention heads &
  12 &
  \textbf{1}\\
number of layers &
  12 &
  12\\
drop rate &
  0.1 &
  0.1\\
query-key-value bias &
  False &
  False\\
total number of parameters &
  162,419,712 &
  \textbf{112,800,768}\\
\tblbottomrule 
\end{tabulary}\par 
\end{table*}
To ensure the fairness of the comparative experiment, both models were trained under exactly the same condition. Firstly, the experiment platform is Kaggle.com powered with a GPU-100 accelerator, and pytorch of version 2.6.0+cu124 is implemented to carry out the experiment. Then, a novel \textit{War and Peace }downloaded from gutenberg.org is adopted as the dataset for this experiment, which contains 3,227,578 characters or 854094 tokens. 90\% of them are used as a training set, and the rest 10\% are used as a validation set. Finally, loss functions are defined based on the cross-entropy function provided by pytorch, and by every 100 batches a training loss and a validation loss are calculated and recorded, then analyzed and discussed in the following section.
    
\section{Results and Discussions}
The experiment runs 5 epochs in total, model a costs 15 minutes and 47 seconds, and model b costs 9 minutes and 51 seconds. This result is exciting, as model b saves 37.6\% of the time which model a costs. Considering the training scale of a real LLM in practice, model b with faster training speed may save millions of dollars and tons of energy. Furthermore, model b is not only faster than model a, but also shows better learning capability than model a as shown in Table~\ref{tw-a6c838e0f78c} and Figure~\ref{f-063b9d0497a6} .

\begin{table*}[!htbp]
\caption{{Training lossess and validation losses of the two models} }
\label{tw-a6c838e0f78c}
\def\arraystretch{1}
\ignorespaces 
\centering 
\begin{tabulary}{\linewidth}{p{\dimexpr.11720000000000002\linewidth-2\tabcolsep}p{\dimexpr.205\linewidth-2\tabcolsep}p{\dimexpr.23639999999999997\linewidth-2\tabcolsep}p{\dimexpr.2083\linewidth-2\tabcolsep}p{\dimexpr.2331\linewidth-2\tabcolsep}}
\tbltoprule Steps & Model a training loss & Model a validation loss & Model b training loss & Model b validation loss\\
\tblmidrule 
0 &
  9.884 &
  9.865 &
  10.621 &
  10.741\\
100 &
  6.055 &
  6.394 &
  6.023 &
  6.309\\
200 &
  5.842 &
  6.019 &
  5.468 &
  5.957\\
300 &
  5.592 &
  5.866 &
  5.423 &
  5.852\\
400 &
  5.405 &
  5.804 &
  5.287 &
  5.730\\
500 &
  5.342 &
  5.686 &
  5.194 &
  5.654\\
600 &
  5.404 &
  5.682 &
  5.051 &
  5.599\\
700 &
  5.491 &
  5.548 &
  5.121 &
  5.555\\
800 &
  5.020 &
  5.526 &
  4.819 &
  5.471\\
900 &
  5.011 &
  5.432 &
  4.788 &
  5.456\\
1000 &
  5.448 &
  5.404 &
  4.788 &
  5.359\\
1100 &
  5.130 &
  5.433 &
  4.575 &
  5.358\\
1200 &
  5.076 &
  5.453 &
  4.819 &
  5.375\\
1300 &
  5.171 &
  5.365 &
  4.849 &
  5.344\\
1400 &
  4.594 &
  5.339 &
  4.785 &
  5.320\\
1500 &
  4.909 &
  5.298 &
  4.572 &
  5.274\\
\tblbottomrule 
\end{tabulary}\par 
\end{table*}

\bgroup
\fixFloatSize{images/c1860237-184d-470d-91a3-cc6a37ebb8ba-ufigure4-uloss-comparation.png}
\begin{figure*}[!htbp]
\centering \makeatletter\IfFileExists{images/c1860237-184d-470d-91a3-cc6a37ebb8ba-ufigure4-uloss-comparation.png}{\includegraphics{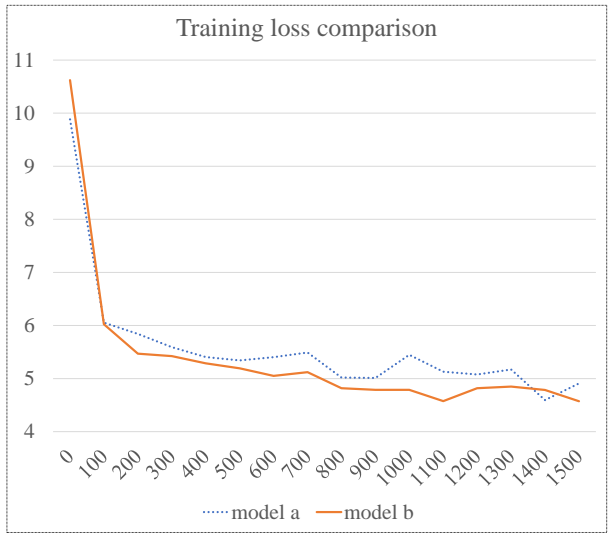}}{\includegraphics{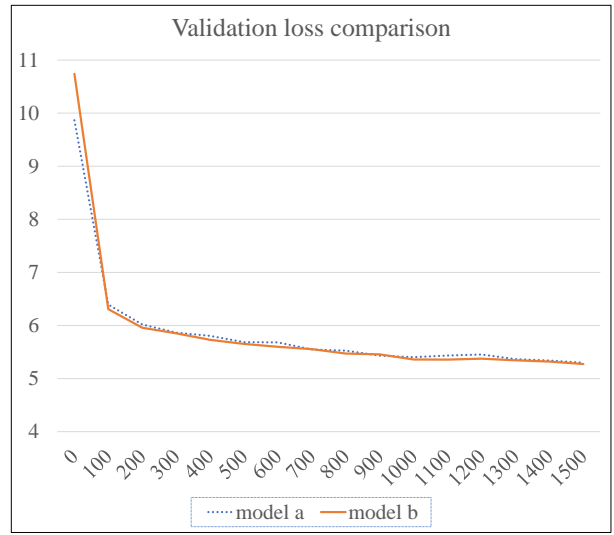}}
\makeatother 
\caption{{Loss comparison of the two models}}
\label{f-063b9d0497a6}
\end{figure*}
\egroup
Table~\ref{tw-a6c838e0f78c} shows the training losses and validation losses of both model a and model b during the 1st epoch of the experiment, and Figure~\ref{f-063b9d0497a6} summarizes the results visually. The rest 4 epochs are ignored here as both the two models have already been close to convergence in the first epoch, and they show obvious signs of overfitting in the subsequent 4 training epochs. 

Overall, model b performs better than model a in terms of learning potential, learning rate, and learning stability, at least no worse than model a. At the last step 1500, the training loss value of model a is 4.909, yet the training loss value of model b is 4.572, and this shows that model b learns deeper than model a. Model a reaches training loss 5.000 at step 1400, yet model b reaches training loss 5.000 at step 800, and this shows that model b learns quicker in each step than model a. The plotting line of model b is smoother than model a, and this means that model b acts more stable than model a in training processes.

In summary, with the single-head DVA block, model b shows better learning capabilities, meanwhile saving a considerable amount of computing resources than model a. This is very valuable in practice, because it shows a high, at least not very low, possibility for us to train efficient LLMs at a low level of cost in time, energy and money.
    
\section{Related Works}
To optimize the architecture of transformer, several works have been proposed. Some works \unskip~\cite{2850382:34235745}  try to mitigate the complexity with static sparse attention mechanisms, in which a static mask matrix is used to filter out the unnecessary scores created by the dot-product of queries and keys. Making the sparse attention dynamic \unskip~\cite{2850382:34235754}  is an improvement of this kind of works, yet it is still constrained by the number of heads.

Another kind of work is called gating mechanism which is derived from LSTM \unskip~\cite{2850382:34216537} . Such works usually control what is remembered and what is forgotten during the information flows of LLMs. They are more sensitive than sparse attention mechanisms, because a mask is composed of only 1 or 0, and a gating mechanism assigns a weight to each input embedding. Originally, gating mechanisms are used in various kinds of RNNs, a recent work \unskip~\cite{2850382:34235756}  tries to plug it into the transformer block. A non-liner function, such as sigmoid function, is used as a gate to decide how much to remember. This work could be interpreted as another kind of dynamic sparse attention by optimizing the score matrix, while still limited by the number of heads.

Mamba \unskip~\cite{2850382:34216538}  tries to compete with transformer by adopting an advanced linear RNN model, and it seems working well at small model sizes. However, as mentioned earlier, it has an intrinsic limitation in paying attention to individual input embeddings separately, and accumulating them into a single hidden layer. 

In summary, historical works do not take care of the semantic relationships between queries and keys, a fixed value is provided for all the queries in a head, and this constrains the learning capability of LLMs. However, DVA creates a specific value update for each pair of query and key, and this enables queries to fetch richer semantics in a single head, meanwhile cutting down the entire FFN block.
    
\section{Conclusion}
This paper provides two main contributions, including a semantic explanation of LLM, and a reconstruction of transformer with DVA. It is worth noting that although the reconstruction of transformer seems valuable, it would have not been achieved without the semantic explanation of transformer proposed in the baseline section. Interpretability has been one of the most major limitations for LLMs for a long time, so that a clearer explanation not only provides a more solid foundation for the field of AI, but also offers various possible ideas for constructions and optimizations of LLMs. The reconstructed model with DVA in this paper just confirms this possibility, and future work could be promising, such as encapsulating DVA into a plug-in module, like LoRA \unskip~\cite{2850382:34234812} , for tuning LLMs in various specific conditions.
\section*{Acknowledgements}The experiment could be found at \BreakURLText{https://www.kaggle.com/xw0220}

\bibliographystyle{elsarticle-num}

\bibliography{\jobname}

\end{document}